\documentclass[10pt,twocolumn,letterpaper]{article}

\usepackage[pagenumbers]{cvpr} %

\usepackage[utf8]{inputenc} %
\usepackage[T1]{fontenc}    %
\usepackage{url}            %
\usepackage{booktabs}       %
\usepackage{amsfonts}       %
\usepackage{amssymb}
\usepackage{amsthm}
\usepackage{algorithm}
\usepackage{algpseudocode}
\usepackage{nicefrac}       %
\usepackage{microtype}      %
\usepackage{xcolor}         %
\usepackage{graphicx}
\usepackage{caption}
\usepackage{subcaption}
\usepackage{mathtools}
\usepackage{siunitx}
\usepackage{svg}
\usepackage{wrapfig}
\usepackage{enumitem}
\usepackage{adjustbox}
\usepackage{enumitem}
\usepackage{commath}
\usepackage{relsize}
\usepackage{tabularx}
\usepackage{bbm}
\usepackage{placeins}
\usepackage[table]{xcolor}
\usepackage[most]{tcolorbox}

\usepackage{takeaway}

\definecolor{cvprblue}{rgb}{0.21,0.49,0.74}
\usepackage[pagebackref,breaklinks,colorlinks,allcolors=cvprblue]{hyperref}

\def\confName{CVPR}
\def\confYear{2027}

\title{FF-JEPA: Action-Free Task-Conditioned Subgoal Learning \\ for Long-Horizon Planning with Latent World Models}

\author{
    Sergi Masip\textsuperscript{1,*},
    Jonathan Swinnen\textsuperscript{1,*},
    Yurui Du\textsuperscript{1,2},
    Renaud Detry\textsuperscript{1,3},
    Tinne Tuytelaars\textsuperscript{1}
    \\
    \small{
        \textsuperscript{*}Equal contribution
        \quad
        \textsuperscript{1}ESAT/PSI, KU Leuven
        \quad
        \textsuperscript{2}PMA, KU Leuven
        \quad
        \textsuperscript{3}RAM, MECH, KU Leuven
    }
    \\
    \small\texttt{\{sergi.masipcabeza, jonathan.swinnen\}@kuleuven.be}
}

\usepackage{graphicx}
\usepackage{caption}

\usepackage{graphicx}

\makeatletter
\apptocmd{\@maketitle}{%
    \begin{center}
        \includegraphics[width=1\textwidth]{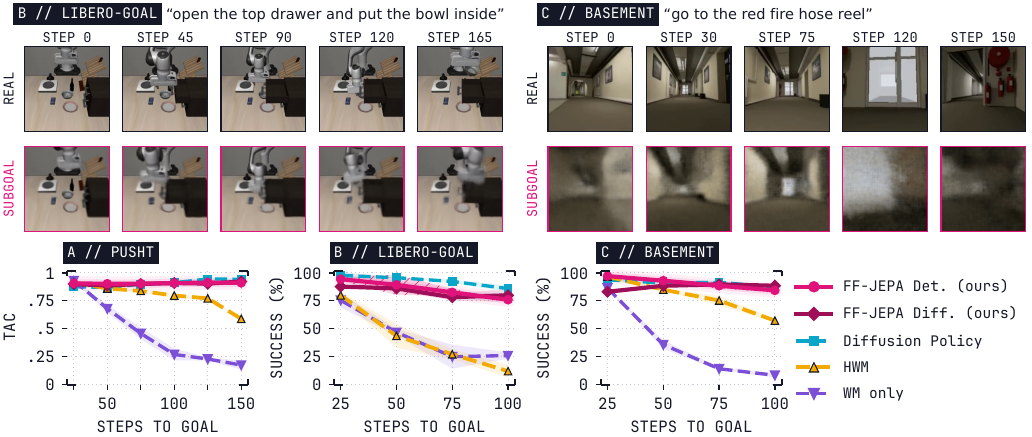}
        \captionof{figure}{%
            \textbf{FF-JEPA} predicts task-conditioned subgoals from action-free demos, guiding a frozen world model without goal images or high-level search. Top: predicted subgoals (decoded for visualisation only). Bottom: Success vs. goal distance. Unlike goal-image planning (WM only, HWM), FF-JEPA remains robust over increasing horizons, comparable to an action-supervised Diffusion Policy.
        }
        \label{fig:teaser}
    \end{center}
    \vspace{3mm}
}{}{\PatchFailed}
\makeatother

\definecolor{Ink}{HTML}{151826}
\definecolor{Ours}{HTML}{E0137A}
\definecolor{OursLight}{HTML}{F8BCD9}   %
\definecolor{Control2}{HTML}{8A8F9E}
\definecolor{Grid}{HTML}{C9CCD6}
 
\newcommand{\sd}[1]{{\scriptsize\color{Control2}$\pm$#1}}
\newcommand{\best}{\cellcolor{OursLight!60}}

\begin{document}
\maketitle

\begin{abstract}
    Latent world models such as Joint-Embedding Predictive Architectures (JEPAs) can plan by searching for actions that reach a goal image, but planning degrades as the horizon grows. We argue this is partly because a goal image shows a task's outcome, not how to get there. We ask whether this search can be replaced by prediction instead: if a world model already captures how actions change the scene, can tasks be learned from action-free data? We explore this with Forward-Forward JEPA (FF-JEPA), which splits the problem into two forward models: (1) a world model trained on interaction data without task labels or rewards, and (2) a subgoal predictor trained on demonstrations without action labels. The predictor proposes the next latent subgoal for the task, and the frozen world model plans towards it with model predictive control, without a goal image or high-level search. On PushT, LIBERO-Goal, and Basement, a new quadruped navigation benchmark, FF-JEPA outperforms flat and hierarchical goal-image planners at long horizons and is comparable to a Diffusion Policy trained with action labels.
\end{abstract}

\section{Introduction}

Learning to act in complex environments requires both understanding how actions change the world and knowing which changes lead to task completion. These two forms of knowledge are often learned together: policies are trained from task demonstrations paired with action labels, directly associating observations and task objectives with the actions needed to achieve them. 
The resulting policy entangles world modeling with task learning. This makes its training difficult, as expert data is scarce. However, noisy or play data is easier to collect and can be used to train a world model that learns the dynamics of the world. This raises a natural question: if an agent already has a world model of how actions change the scene, does learning a new task still require action-labeled demonstrations? %

Action-free learning suggests a positive answer. Rather than directly mapping observations to actions, predictive inverse dynamics models (PIDMs)~\citep{xie2025latent, schafer2026pidm} have shown that combining an inverse dynamics model with an action-free forward model has enabled training on unlabeled trajectories. 
From a vision-language-action perspective, \citet{zhang2026disentangled} separate the pretraining of forward and inverse dynamics models to improve representation learning and downstream control. These perspectives suggest planning over predicted future states rather than directly mapping observations to actions.

In recent work, Joint-Embedding Predictive Architecture (JEPA) \cite{lecun2022path, assran2023ijepa} world models have been shown to be a promising approach for planning the actions between different states. These models enable short-horizon planning by imagining actions and their consequences and optimizing them with model predictive control (MPC)~\citep{zhou2025dinowm, maes2026lewm}, for example, using the Cross-Entropy Method (CEM)~\citep{rubinstein2004cross}. In reward-free settings, tasks are commonly specified through a goal observation, and the planner searches for actions that bring its predicted latent state closer to that goal~\citep{zhou2025dinowm, maes2026lewm}. Yet a goal image specifies the desired outcome, not how the task should progress from the current state. As the distance to the goal increases, planning becomes increasingly difficult. Prediction errors accumulate over longer rollouts, and the planner must discover a path connecting the current observation to the desired outcome. Hierarchical approaches introduce intermediate subgoals~\citep{zhang2026hwm, zhang2026hjepa, zhou2026idwm}, but generally remain guided by a goal observation. They improve how the planner searches for a specified goal, but do not learn how a more complex task should progress.

We therefore ask: can we use a pretrained latent world model to provide the action knowledge for action-free task learning? Rather than training a model to predict the actions directly, we explore using planning with a world model as the mechanism that turns predicted task progress into actions. 
This separates the two complementary forms of knowledge: how actions change the environment and how task progress unfolds. The former can be learned in a task-agnostic manner from interaction data without task labels, rewards, or success flags, while the latter can be acquired from task demonstrations without action labels.

We explore this idea with a simple approach, Forward-Forward JEPA (FF-JEPA)%
, which pairs a frozen latent world model with an action-free, task-conditioned subgoal predictor. The world model is trained on interaction data containing observations and actions, but no task labels, rewards, or success annotations. The subgoal predictor is trained on a small set of expert demonstrations paired with task descriptions, but without action labels. At inference time, given only the current observation and task, the predictor proposes future latent subgoals that describe how the task should progress from the current state. The world model then searches for actions towards these subgoals using CEM. By repeatedly searching towards predicted subgoals, FF-JEPA decomposes long-horizon planning into a sequence of short-horizon search problems, without requiring a goal image or high-level search.

We evaluate FF-JEPA on three environments spanning planar pushing (PushT \cite{chi2025dp}), tabletop manipulation (LIBERO-Goal \cite{liu2023libero}), and quadruped navigation (Basement). Across these settings, FF-JEPA outperforms flat and hierarchical goal-image world-model planners, particularly as the distance to the goal increases, and achieves comparable performance to that of a Diffusion Policy \cite{chi2025dp} trained on the same demonstrations with action labels. We further investigate how world-model representations affect downstream planning, how deterministic and stochastic subgoal predictors behave across tasks, and how performance changes with the number of available demonstrations. We summarize our contributions as follows:

\begin{itemize}
  \item \textbf{A frozen latent world model as the action side.} 
  We study whether
  tasks learned from action-free demonstrations can be executed by a world model trained
  on noisy interaction data without task labels, rewards, or success flags. We instantiate this with FF-JEPA, which needs no goal image and no high-level search.
  \item \textbf{Long-horizon planning without goal images or action-labeled demonstrations.}
  On PushT, LIBERO-Goal and Basement, a new quadruped 3D navigation benchmark, FF-JEPA outperforms flat and hierarchical goal-image planners, stays robust on longer trajectories, and compares to a Diffusion Policy trained with action labels.
  \item \textbf{What the world model must encode.}
  The best world model encodes object state in manipulation in PushT and LIBERO-Goal, and monotonic progress in our navigation environment. Monotonicity matters less when subgoals provide the long-range direction. 
  \item \textbf{A subgoal predictor learnable from a handful of demonstrations.} While not as powerful as a fully trained model, it can learn even with as few as 1--7 demonstrations per task in LIBERO-Goal.
\end{itemize}

\section{Related Work}\label{sec:related}

\paragraph{Latent world models for planning.}

Latent world models learn environment dynamics and representations and use them for decision-making. In model-based RL, actions are selected by planning over predicted rewards~\citep{hafner2019planet,hansen2024tdmpc2} or by optimizing a policy inside the learned model~\citep{ha2018worldmodels,hafner2025dreamerv3}. Reward-free approaches specify tasks as goal observations instead, and act by optimizing action sequences that minimize the distance between predicted and goal embeddings~\citep{zhou2025dinowm,assran2025vjepa2,sobal2025pldm}. This is typically done using CEM~\citep{rubinstein2004cross} or MPPI~\citep{williams2017mppi}, which can also be further augmented with a learned proposal distribution to make planning more reliable~\citep{wang2026prism,chahe2026pijepa}. Such a planning procedure can be applied to generative world models that predict future frames in pixel space, for example, for visual navigation~\citep{bar2025nwm}. In contrast, latent world models, such as Joint-Embedding Predictive Architectures (JEPAs)~\citep{lecun2022path,assran2023ijepa}, enable planning over lower-dimensional latent states. Their representation can either come from a frozen pretrained encoder~\citep{zhou2025dinowm,assran2025vjepa2,terver2025whatdrives} or be learned end-to-end together with the world model~\citep{sobal2025pldm,maes2026lewm}. How this representation is trained matters, since it can collapse to a trivial solution. To avoid this, anti-collapse techniques such as VICReg~\citep{bardes2022vicreg}, SIGReg~\citep{balestriero2025lejepa} or spectral alignment~\citep{molin2026specwm}, or additional inverse dynamics objectives~\citep{brandfonbrener2023inverse} are used. Despite these advances, long-horizon planning remains challenging. Goal-conditioned approaches typically require a goal image at evaluation, which can be unavailable in real deployments. Moreover, in this situation, the planner has to bridge from the current state to a distant goal, a difficult task when nothing informs about the route. FF-JEPA addresses both limitations by decomposing tasks into a sequence of predicted subgoals, allowing it to solve long-horizon problems without an explicit goal image.

\paragraph{Hierarchical world models.}
Hierarchical RL agents select latent goals or abstract states inside a reward-trained world model~\citep{hafner2022director,gumbsch2024thick}, and hierarchical visual foresight~\citep{nair2020hierarchical}  optimizes image subgoals in a generative model's latent space.
In reward-free latent world models, HWM~\citep{zhang2026hwm} and Hi-LeWM~\citep{caselli2026mindthegap} search over learned latent macro-actions. Concurrently, H-JEPA~\citep{zhang2026hjepa} trains a stack of action-conditioned JEPAs end-to-end, and ID-WM~\citep{zhou2026idwm} searches the noise of an action-free generator of reachable states. 
Another work uses latent flow towards the goal~\citep{huang2026leflow}. These approaches need a reward or a goal image to specify the task. FF-JEPA predicts task-conditioned subgoals, with a predictor trained on a few demonstrations on top of a task-free world model, so it needs no goal image or reward at test time.

\paragraph{Subgoal generation for control.}

Text-conditioned video or image generators can be used to produce visual plans or subgoals. These are then executed by inverse dynamics or goal-conditioned policies~\citep{du2023unipi,du2024vlp,ajay2023hip,black2024susie}. Video-prediction policies use predicted futures as features for an action head~\citep{hu2025vpp,tian2025seer}, and \citet{schafer2026pidm} analyse when such predictive inverse dynamics outperform behaviour cloning. However, pixel generation is costly, and the action model is often task-specific. Latent-action models instead infer actions from unlabelled video~\citep{schmidt2024lapo,ye2025lapa}. Other approaches plan in state or latent space and use inverse dynamics or goal-conditioned policies for execution~\citep{ajay2023is,park2023hiql}. Demo-JEPA~\citep{he2026demojepa} also uses demonstration-derived latent subgoals for planning in a JEPA world model, but targets one-shot cross-embodiment imitation. %
Concurrent work generates subgoals that both condition an action prior and serve as planning targets~\citep{cheng2026sage}, or jointly learns dynamics and language-conditioned subgoals with a single shared predictor~\citep{barbeau2026lago}. FF-JEPA trains its task-conditioned subgoal predictor on top of a frozen, task-free world model, so it works across world models and learns new tasks from a few demonstrations without retraining the dynamics. LDP~\citep{xie2025latent} predicts dense sequences of action-free latents in a frozen VAE space and recovers actions with an inverse-dynamics model. FF-JEPA instead reaches sparse subgoals with an action-conditioned world model.

\paragraph{Decoupling world modeling from task supervision.}
Learning from play~\citep{lynch2019play,lynch2021mcil,mees2022hulc} trains a single goal- or language-conditioned policy on action-labeled play data without task labels, with language attached in hindsight to under 1\% of it~\citep{lynch2021mcil,mees2022hulc}. World models also steer, rank, or co-train policies~\citep{zhu2025uwm,qi2025gpc,wang2025lps}. Forward and inverse dynamics can be pretrained separately before policy learning~\citep{zhang2026disentangled}, but the policy, trained on action-labeled task data, still produces the actions, as in Diffusion Policy and VLAs~\citep{chi2025dp,kim2024openvla,black2024pi0}.
FF-JEPA splits the two: task knowledge lives in an action-free subgoal predictor trained on a few task demonstrations, and action knowledge lives in a task-free world model trained without task labels, so the subgoal predictor can be retrained independently.

\begin{figure*}[ht]
    \centering
    \includegraphics{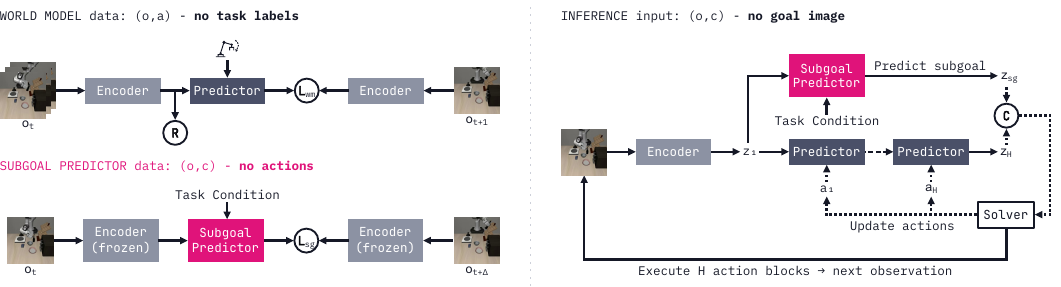} 
    \caption{\textbf{FF-JEPA.} A subgoal predictor is trained on action-free demos to generate task-conditioned subgoals in the frozen world model's latent space. At inference, the world model is used to find the actions that lead to each subgoal through CEM.}\label{fig:method}
\end{figure*} 
\section{Forward-Forward JEPA (FF-JEPA)}

We hypothesize that an agent utilizing a world model to solve tasks needs two kinds of knowledge that can be learned from separate data: how actions change the environment, and where the task goes next. We keep them in separate models that never see the other’s supervision. The world model (\cref{sec:wm}) is trained only on interaction data with actions and no task labels. These trajectories need not be expert and may be noisy or exploratory. On the other hand, the subgoal predictor (\cref{sec:sgp}) is trained only on a few expert task demonstrations without actions. Planning is the only place they meet (\cref{sec:inference}). Every design choice below follows from keeping this separation strict and the connection between them minimal.

\subsection{Problem setting}
\label{sec:setting}
We consider two datasets with complementary supervision. The \emph{interaction dataset} $\mathcal{D}_\text{wm}$ %
contains trajectories in the form of video frames $\mathbf{o}_t$ paired with actions $\mathbf{a}_t$ but without task labels, rewards, or success annotations. It may be noisy, exploratory, or collected for other tasks. The \emph{task dataset} $\mathcal{D}_\text{task}$ %
contains a small number of demonstration videos, each paired with a task descriptor $c$ (a language instruction in our multi-task benchmarks), and has \textbf{no action labels}. At test time, the agent receives the current observed frame $\mathbf{o}_t$ and task $c$, but no goal image, and must output actions to solve $c$. Tasks are drawn from a fixed set of instructions. We do not study generalization to unseen instructions.

From $\mathcal{D}_{\text{wm}}$ we learn a latent world model consisting of two components: an encoder $E$ that maps an observation to a latent state $\mathbf{z}_t = E(\mathbf{o}_t)$, and an action-conditioned predictor $P$ that predicts the next latent state from a window of $W$ past states and actions,
\begin{equation}
  \hat{\mathbf{z}}_{t+1} = P\big(\mathbf{z}_{t-W+1:t},\, \mathbf{a}_{t-W+1:t}\big).
  \label{eq:predictor}
\end{equation}
Actions $\mathbf{a}_{t}$ are grouped into blocks of $k$ environment steps, and $t$ indexes these blocks. In Secs.~\ref{sec:inference} and~\ref{sec:sgp} we treat $E$ and $P$ as given and frozen; Sec.~\ref{sec:wm} describes how they are obtained.

\subsection{From world modeling to task completion}
\label{sec:inference}
The world model understands the dynamics of the scene and can reliably search towards nearby states. Given a target latent $\mathbf{z}^\star$, the world model acts by planning a horizon of $H$ action blocks
\begin{equation}
  \hat{\mathbf{a}}_{t:t+H-1} = \arg\min_{\mathbf{a}_{t:t+H-1}}
  \big\| P_{\text{AR}}(\mathbf{z}_t, \mathbf{a}_{t:t+H-1}) - \mathbf{z}^\star \big\|_2^2,
  \label{eq:plan}
\end{equation}
where $P_{\text{AR}}$ rolls out $P$ autoregressively over an extended sequence of action blocks. We solve Eq.~\eqref{eq:plan} using CEM~\cite{rubinstein2004cross}. Because latent world models can be sensitive to out-of-distribution transitions, we initialize CEM from a task-free action prior $\pi(\mathbf{a}_t \mid \mathbf{z}_t)$. This way, we keep actions in the distribution the world model was trained on~\cite{wang2026prism}. Usually, $\mathbf{z}^\star = E(\mathbf{o}_g)$ encodes a goal image~\cite{maes2026lewm,zhou2025dinowm}. However, images that portray the outcome of a task are often not available at deployment time, and solving a task often implies searching over a long horizon and understanding the task structure. 

We believe that search towards a goal image might not be enough. Therefore, we introduce a subgoal predictor that learns the task from action-free demos and can guide the world model during planning. The subgoal predictor $G$ proposes the next $N$ latents spaced $\Delta$ blocks apart:
\begin{equation}
    \begin{aligned}
    G(\mathbf{z}_{t+\Delta}, ..., \mathbf{z}_{t+N\Delta} \mid \mathbf{z}_t, c)
    &\approx p_{\mathcal{D}_\text{task}}(\mathbf{z}_{t+\Delta}, ..., \mathbf{z}_{t+N\Delta} \mid \mathbf{z}_t,c) \\
    \hat{\mathbf{g}}_{1:N}
    &\sim G(\,\cdot \mid \mathbf{z}_t, c)
    \end{aligned}
  \label{eq:subgoals}
\end{equation}

Every $H = \Delta$ blocks, the agent samples $S$ subgoal sequences, plans towards the first subgoal of each with batched CEM, executes the plan with the lowest final cost, and re-plans from the state reached (Alg.~\ref{alg:inference}, Fig.~\ref{fig:method}). $G$ knows the steps needed to solve a task, but does not know how to reach them. $P$ knows how to execute a step, but does not know what the next step should be. A long-horizon task is thus decomposed into a sequence of short search problems, each within the range where latent planning is reliable.

\begin{algorithm}[t]
\caption{FF-JEPA inference}\label{alg:inference}
\label{alg:ffjepa}
\begin{algorithmic}[1]
\Require task $c$; world model $E,P$; action prior $\pi$; subgoal predictor $G$; horizon $H$; samples $S$
\While{task not solved and budget not exhausted}
  \State $\mathbf{z}_t \gets E(\mathbf{o}_t)$
  \State $\hat{\mathbf{g}}^{(s)}_{1:N}\sim G(\cdot\mid\mathbf{z}_t,c)$ for $s=1,\dots,S$
  \For{$s=1,\dots,S$}
    \State $(\hat{\mathbf{a}}^{(s)}_{t:t+H-1}, J^{(s)})\gets\textsc{CEM}\big(\ \mathbf{z}^\star{=}\hat{\mathbf{g}}^{(s)}_1,\ \text{init}\sim\pi\big)$
  \EndFor
  \State $s^\star\gets\arg\min_s J^{(s)}$ \Comment{lowest planning cost}
  \State execute $\hat{\mathbf{a}}^{(s^\star)}_{t:t+H-1}$ ($Hk$ environment steps)
\EndWhile
\end{algorithmic}
\end{algorithm}

\subsection{Task-conditioned subgoal predictor}
\label{sec:sgp}
$G$ models task progress: where demonstrations of task $c$ go next from state $\mathbf{z}_t$. It is the second forward model in FF-JEPA: $P$ predicts the effect of an action, $G$ the effect of pursuing the task. Unlike hierarchical world models~\cite{cheng2026sage,zhang2026hwm,zhou2026idwm}, it needs no goal observation and no search.

We train $G$ on windows $(\mathbf{z}_t;\, \mathbf{z}_{t+\Delta}, \dots, \mathbf{z}_{t+N\Delta})$ of demonstrations encoded by the frozen encoder $E$ with their task descriptor $c$ (e.g., frozen CLIP~\cite{radford2021learning} text tokens in our experiments) and no actions. Subgoals are sampled $\Delta$
action blocks apart along feasible trajectories, so each subgoal is reachable. Windows past the end of a demonstration are padded with its final latent, so $G$ learns to stay at the goal.
Since long-horizon tasks can often be solved via several alternative trajectories, we compare two planners, which differ in how they represent the distribution of solutions.

\paragraph{Deterministic planner $G_{\text{Det.}}$} predicts the \emph{expected} next subgoal. It is trained with MSE and outputs the mean of the conditional distribution of potential next subgoals. We can thus only sample a single proposed trajectory ($S=1$). This suits tasks where we assume the distribution of subgoals is unimodal, but it might yield states that are not reachable or desirable for tasks where the distribution has multiple modes. It is a causal transformer \cite{vaswani2017attn} over $c$ followed by $\mathbf{z}_t, \hat{\mathbf{g}}_1, \hat{\mathbf{g}}_2, \dots$
 
\paragraph{Diffusion planner $G_{\text{Diff.}}$} models the entire conditional \emph{distribution} of next subgoals. We can use it to sample $S$ paths from which the world model can select the most feasible one. A transformer denoises the $N$ subgoals jointly, cross-attending to $\mathbf{z}_t$ and the task condition. It is trained with v-prediction~\cite{salimans2022progressive} and sampled with DDIM~\cite{song2021denoising}. It is computationally more expensive than $G_{\text{Det.}}$, but does not require the assumption that the subgoal distribution is unimodal.

\subsection{Latent world model}
\label{sec:wm}
FF-JEPA works with any JEPA world model whose latent distances are meaningful, since they serve both as the planning cost and as the space $G$ predicts in. In our experiments, $E$ is a ViT~\cite{dosovitskiy2021vit}, either trained from scratch or from a frozen pretrained model, whose patch tokens are pooled by a CLS token or $Q$ learned queries~\cite{jaegle2021perceiver}; $P$ is a causal transformer~\cite{maes2026lewm, zhou2025dinowm}. Both are trained on $\mathcal{D}_{\text{wm}}$ with
\begin{equation}
  \mathcal{L}_{\text{wm}} = \big\| \hat{\mathbf{z}}_{t+1} - \mathbf{z}_{t+1} \big\|_2^2
  + \lambda\, R(\mathbf{z}, \mathbf{a}),
  \label{eq:wm}
\end{equation}
where $R$ is an objective that prevents collapse which, in doing so, determines what the latent represents. We consider four choices: SIGReg~\cite{balestriero2025lejepa,maes2026lewm} spreads information evenly but has no preference for action-relevant content; SpecWM~\cite{molin2026specwm} favours temporally smooth latents whose distances track elapsed time; inverse dynamics~\citep{brandfonbrener2023inverse} forces the latent to encode action effects; and a frozen DINO encoder with $R = 0$~\cite{zhou2025dinowm} offers generic features not tuned to the dynamics. Since no choice is best at every benchmark, we select $R$ per benchmark by goal-image planning success on held-out $\mathcal{D}_{\text{wm}}$ episodes. The action prior $\pi$ is a Gaussian fitted by maximum likelihood on $\mathcal{D}_{\text{wm}}$. %

\section{Experiments}
\label{sec:experiments}

We evaluate whether FF-JEPA enables long-horizon planning in a variety of environments and tasks, including planar pushing (PushT), manipulation (LIBERO-Goal), and 3D navigation (Basement) (\cref{sec:main_results}). We also investigate how each latent world model variant affects planning on each task (\cref{sec:wm_analysis}) and how deterministic or stochastic subgoals suit each task better (\cref{sec:detsto}). Finally, we analyze data efficiency in the subgoal predictor and discuss its inference time (\cref{sec:efficiency}).

\subsection{Experimental setup}
\label{sec:setup}

\begin{table}[t]
\centering
\footnotesize
\setlength{\tabcolsep}{3pt}
\caption{\textbf{Benchmarks and FF-JEPA configuration.} The world model and the subgoal predictor use different data. The subgoal predictor never uses actions. WM objective selected per benchmark (Sec.~\ref{sec:setup}). Steps are environment steps; success counts at any step.}
\label{tab:benchmarks}
\begin{tabular}{@{}lccc@{}}
\toprule
 & PushT & LIBERO-Goal & Basement \\
\midrule
WM data & 17k rollouts & 20k rollouts & 10k trajectories \\
SP demos (held out) & 1,681 (187) & 378 (50) & 1,800 (200) \\
Task input $c$ & -- & text (10 tasks) & text (20 goals) \\
Cameras & 1 & 2 & 1 \\
WM objective & SIGReg & inverse dyn. & SpecWM \\
$\Delta{=}H$ & 5 & 3 & 3 \\
Eval.\ starts & 256 & 200 & 300 \\
Budget (steps) & 300 & 300 & 600 \\
Success criterion & TAC (soft score) & task check & 0.75\,m, 45$^\circ$ \\
\bottomrule
\end{tabular}
\end{table}

\paragraph{Benchmarks.}
Tab.~\ref{tab:benchmarks} summarizes the three benchmarks. PushT~\citep{chi2025dp} is a single-task setup, where a pusher moves a T-shaped block onto a fixed target. In LIBERO-Goal~\cite{liu2023libero}, a Franka arm solves 10 tasks in one kitchen scene, each specified by an instruction. The world model is trained on rollouts of SmolVLA~\cite{shukor2025smolvla} with added noise and without their task labels. The subgoal predictor is trained on LIBERO's human demonstrations. We also introduce the Basement benchmark, in which a simulated Unitree Go2 quadruped navigates a 3D reconstruction of a real building and must reach one of 20 objects named in an instruction (App.~\ref{app:basement}). Its world-model data mixes goal-seeking, exploratory, and random behavior. In the multi-task benchmarks, the task descriptor $c$ is one of a fixed set of instructions (10 in LIBERO-Goal, 20 in Basement). We use it as a task identifier and do not evaluate generalization to unseen instructions. All splits are at the episode level. We use action blocks of $k=5$ environment steps for all world models in all benchmarks. %

\paragraph{Evaluation.}
We report the target area coverage (TAC)~\citep{chi2025dp} in PushT and task success rate (SR) in LIBERO-Goal and Basement. Main results start from random initial states. The set of initial states is fixed, so all methods start from the same states. LIBERO's initial states come without a goal image, so goal-image methods are evaluated there using the 50 held-out demonstrations instead, with that demonstration's final frame as the
goal (marked $^\ast$). These numbers therefore come from different start states and are not directly
comparable to the other rows. To measure robustness to the distance to the goal (Fig.~\ref{fig:teaser}, bottom), we additionally start every method $d$ steps before the end of a held-out demonstration with a budget of $2d$ steps. This allows methods needing a goal image to use the demonstration's final frame.
We report the mean and standard deviation across two random seeds for all experiments, unless stated otherwise.

\paragraph{FF-JEPA.}
For every world model objective of \cref{sec:wm_analysis}, we measure goal-image planning success on 25-step
windows of held-out $\mathcal{D}_{\text{wm}}$ episodes and keep the objective with the highest success
(Tab.~\ref{tab:benchmarks}). This selection uses no task data. Sec.~\ref{sec:wm_analysis} compares all
objectives. The subgoal predictor predicts $N{=}3$ subgoals, spaced $\Delta$ blocks apart, and the world model
plans towards only the first of them before re-planning, over a horizon $H{=}\Delta$. 
We report both subgoal predictors: deterministic and diffusion (20 DDIM steps,
$S{=}4$ sampled subgoal sequences). Full implementation details and code will be released in a future version. %

\paragraph{Baselines.}
Each baseline receives information that FF-JEPA does not (Tab.~\ref{tab:main}).
\emph{Diffusion Policy} (DP)~\cite{chi2025dp} is trained on the same task demonstrations as the
subgoal predictor, but with action labels (details in App.~\ref{app:baselines}). It does not use $\mathcal{D}_{\text{wm}}$, whose trajectories may be noisy, exploratory, or collected
for other tasks---useful for learning dynamics, but not suitable for imitation.
\emph{WM only} plans with CEM towards a goal image using the same frozen world model, following HWM's
flat-planning protocol~\cite{zhang2026hwm}. \emph{HWM} adds a
high-level predictor over learned macro-actions, trained on the same frozen world model.
These two receive a goal image but no task demonstrations. 
All world-model methods share the same action prior and step budget.

\subsection{Long-horizon planning with FF-JEPA}
\label{sec:main_results}

\begin{table}[t]\centering\setlength{\tabcolsep}{4pt}
\caption{\textbf{Main results}. $^\dagger$: needs a goal image. $^*$: evaluated on the 50 reserved demos as full-task windows.}
\label{tab:main}
\begin{tabular}{@{}lccc@{}}
\toprule
 & PushT & LIBERO-Goal & Basement \\
 & {\scriptsize TAC} & {\scriptsize success (\%)} & {\scriptsize success (\%)} \\
\midrule
Diffusion Policy & \textbf{0.86}\sd{0.01} & \textbf{94.50}\sd{1.41} & 65.17\sd{2.12} \\
\midrule
WM only$^\dagger$ & 0.15\sd{0.03} & \phantom{0}4.26$^*$\sd{3.01} & \phantom{0}5.33\sd{2.36} \\
HWM$^\dagger$ & 0.41\sd{0.01} & \phantom{0}5.32$^*$\sd{1.50} & 29.33\sd{0.47} \\
\midrule
FF-JEPA Det. & \underline{0.82\sd{0.01}} & \underline{89.00\sd{0.71}} & \underline{82.00\sd{2.36}} \\
FF-JEPA Diff. & 0.76\sd{0.01} & 82.50\sd{2.83} & \textbf{90.67}\sd{0.00} \\
\bottomrule
\end{tabular}
\end{table}

In our first experiments, we evaluate FF-JEPA on the original benchmark tasks end-to-end. Table \ref{tab:main} reports its accuracy metrics compared with the baselines. Additionally, in Figure \ref{fig:teaser} (bottom), we measure robustness to the distance to the goal by evaluating the model on windows at different start step distances $d$ before the end of held-out demonstrations, with a budget of $2 \cdot d$ steps.
From these results, we can clearly see that goal-image baselines tend to deteriorate when the planning horizon increases. This is not the case for FF-JEPA, which shows stable performance, regardless of distance to the goal. Its performance only deteriorates somewhat in the full evaluation for Push-T (reported in Table \ref{tab:main}), which uses more extreme, random episode initializations.

\begin{takeaway}
Given a pretrained world model, an action-free subgoal predictor trained on expert demos enables goal- and high-level-search-free long-horizon planning, achieving performance comparable to that of a diffusion policy trained using action labels on the same demos.
\end{takeaway}

\subsection{Which world model suits which task?}
\label{sec:wm_analysis}

\begin{figure}
    \centering
    \includegraphics[width=1\linewidth]{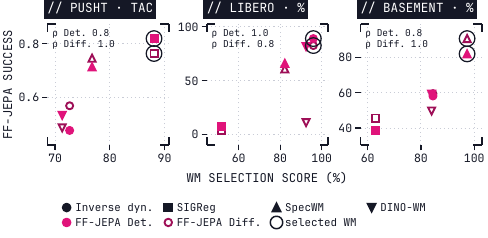}
    \caption{\textbf{Selecting the world model.} Selection score (goal-image planning on held-out $D_\text{wm}$ windows) vs. FF-JEPA success. The world model with best selection score wins on every benchmark for both predictors. No objective wins everywhere.
    }
    \label{fig:wm_ablation}
\end{figure}

We also investigated how the choice of the world model determines the planning performance of FF-JEPA. To address this question, we trained FF-JEPA with four recent latent world model variants in all environments. \Cref{fig:wm_ablation} reports the world model's short-horizon control performance on 25-step windows (Sel.), as in \citet{maes_lld2026swm}, alongside task performance for FF-JEPA variants (Det. and Diff.). The results suggest that \textbf{short-horizon control performance correlates with task performance with FF-JEPA, making it a good proxy for world model in our experiments}. To compute the main results (\cref{sec:main_results}), we selected the world model following this proxy.

\begin{table}[t]
\centering
\footnotesize
\setlength{\tabcolsep}{2.2pt}
\caption{\textbf{Why a different world model wins on each benchmark.} 
    Obj.: object information beyond the agent ($R^2$). Mono.: Spearman $\rho$ between latent distance to the goal and remaining steps. Succ.: FF-JEPA Det. \colorbox{OursLight!60}{Shaded}: selected world model. Details in App.~\ref{app:probes}.
}
\label{tab:probes}
\begin{tabular}{lcccccccc}
\toprule
 & \multicolumn{3}{c}{PushT} & \multicolumn{3}{c}{LIBERO-Goal} & \multicolumn{2}{c}{Basement} \\
\cmidrule(lr){2-4}\cmidrule(lr){5-7}\cmidrule(lr){8-9}
 & Obj. & Mono. & Succ. & Obj. & Mono. & Succ. & Mono. & Succ. \\
\midrule
Inverse dyn. & 0.63 & 0.70 & .48 & \best\textbf{0.57} & \best 0.67 & \best\textbf{89.0} & 0.47 & 58.2 \\
SIGReg & \best\textbf{0.97} & \best 0.33 & \best\textbf{.82} & $-$0.61 & 0.67 & 6.8 & 0.63 & 38.7 \\
SpecWM & 0.54 & \textbf{0.80} & .71 & 0.42 & \textbf{0.91} & 66.0 & \best\textbf{0.80} & \best\textbf{82.0} \\
DINO-WM & 0.78 & 0.79 & .53 & 0.45 & 0.85 & 81.5 & 0.43 & 59.2 \\
\bottomrule
\end{tabular}
\end{table}

Since a different world model was best in each environment, we further analyzed the reasons with a set of probes on held-out expert demonstrations (full analysis in App.~\ref{app:probes}). We report the two that correlated most with success: object state beyond the agent ($R^2$ of a linear probe predicting states of other objects from the latent) and monotonicity (Spearman correlation between the latent distance to the episode's final frame and the remaining steps). 

As shown in \Cref{tab:probes}, in Basement, which has no objects to manipulate, strong monotonicity seems to indicate success. SpecWM tracks progress best and is the best world model. In PushT and LIBERO-Goal, where interactions with an object are central to the task, success follows object information instead. It orders the four world models in LIBERO-Goal. In PushT, the best world model, SIGReg, encodes the T's orientation far better than the rest. Monotonicity is not needed there, since the best world models on both benchmarks have among the least monotonic latent distance ($\rho$ 0.33 and 0.67). We hypothesize that this may be because the subgoal predictor provides the long-range direction and CEM only needs a locally correct distance to the next subgoal.

\begin{takeaway}
   Short-horizon goal-image control is a task-free proxy that selects the best world model for all three benchmarks. The best world model encoded the task-relevant state: the manipulated object in PushT and LIBERO-Goal, and progress along the path in Basement.
\end{takeaway}

\subsection{Deterministic or stochastic subgoals?}
\label{sec:detsto}

\begin{table}[t]\centering\setlength{\tabcolsep}{2pt}
\caption{\textbf{Subgoal predictor ablation per benchmark.} $S$ sampled subgoal sequences. $^\diamond$: default.
}
\label{tab:abl2_all}
\begin{tabular}{@{}lccc@{}}
\toprule
 & PushT & LIBERO-Goal & Basement \\
 & {\scriptsize TAC} & {\scriptsize success (\%)} & {\scriptsize success (\%)} \\
\midrule
Det. & \textbf{0.82}\sd{0.01} & \textbf{89.00}\sd{0.71} & 82.00\sd{2.36} \\
Diffusion, $S{=}1$ & 0.77\sd{0.00} & 79.00\sd{4.24} & \textbf{91.83}\sd{0.24} \\
Diffusion, $S{=}4^\diamond$ & 0.76\sd{0.01} & 82.50\sd{2.83} & 90.67\sd{0.00} \\
Diffusion, $S{=}8$ & 0.77\sd{0.03} & 84.50\sd{1.41} & 89.83\sd{0.71} \\
\bottomrule
\end{tabular}
\end{table}

\begin{figure}
    \centering
    \includegraphics[width=1\linewidth]{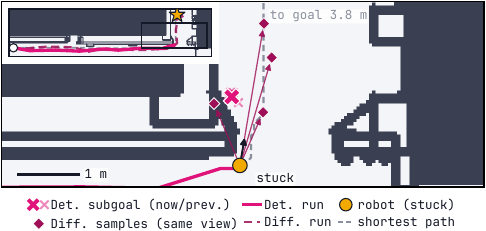}
    \caption{\textbf{Deterministic subgoals can get stuck.} Basement episode where Det. gets stuck on a corner. Det. keeps predicting the same subgoal, while Diff. samples diverse subgoals, most of them leading into the corridor. Positions from a learned position probe.}
    \label{fig:placeholder}
\end{figure}

We report results for two subgoal predictor architectures in Table \ref{tab:main}. Now, we analyze why and in which situations a deterministic or a stochastic diffusion predictor works best. A diffusion architecture should help when the task to solve is multimodal. Recent work has found that the demonstrations of LIBERO and PushT are nearly unimodal, and deterministic regression is competitive on them~\citep{mazza2026multimodality}. The results in \cref{tab:abl2_all} are consistent with these observations, as the deterministic predictor works best in PushT and LIBERO-Goal. Diffusion yields the highest success rate in Basement, but this is not necessarily explained by multimodality, since the deterministic predictor also achieves a good success rate. 

Further analysis reveals that most failures with the deterministic predictor in Basement occur when the robot gets stuck against a wall (see App.~\ref{app:qualitative}). When this happens, the deterministic predictor keeps predicting the same subgoal indefinitely, keeping the robot stuck. In contrast, the diffusion predictor samples different subgoals at each replan until one is eventually sampled that allows the robot to free itself, as illustrated in \cref{fig:placeholder}. Diffusion with $S=1$ performs on par with $S=4$ and $S=8$ on Basement, suggesting that the benefit comes from re-sampling subgoals across re-plans. In the other environments, however, this stochasticity seemed to harm planning rather than help. Therefore, a simple deterministic predictor might often be sufficient in many standard robotics benchmarks, especially given that it also introduces less overhead.

\begin{takeaway}
    The assumption that the conditional subgoal distributions are roughly unimodal is a reasonable approximation in the tested environments. Sometimes, though, stochasticity may be useful to get out of stuck positions, but in other cases it may underperform.
\end{takeaway}

\subsection{Compute and data efficiency}
\label{sec:efficiency}

\begin{figure}
    \centering
    \includegraphics[width=1\linewidth]{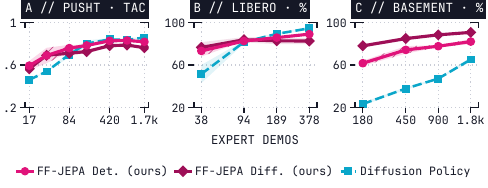}
    \caption{\textbf{Data efficiency.} Performance vs. number of training demonstrations, with gradient steps held fixed.}
    \label{fig:data_efficiency}
\end{figure}

\begin{figure}
    \centering
    \includegraphics[width=1\linewidth]{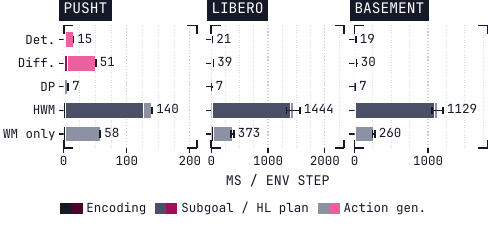}
    \caption{\textbf{Inference cost.} Wall-clock time per environment step on one NVIDIA RTX 3080, split into observation encoding, subgoal / high-level planning, and action generation.}
    \label{fig:inference_cost}
\end{figure}

Finally, we investigated the data efficiency of our approach. First, we evaluated how much training data is required to learn the subgoal predictors. For each benchmark, we train predictors on progressively smaller subsets of the episodes in $\mathcal{D}_\text{task}$, while keeping the total number of gradient steps fixed. We also included diffusion policy trained on the same demonstrations with actions for reference. The results are shown in Figure \ref{fig:data_efficiency}. 

The subgoal predictors remain robust even under few-demonstration regimes. Note that LIBERO and Basement already contain relatively few expert trajectories per task, with an average of only 38 and 90 demonstrations per task, respectively. Even with a random training subset of only 10\% of their expert demonstrations, which amounts to just 1–7 demonstrations per task for LIBERO and 2–16 for Basement, the models still retain a considerable degree of task-solving capability. These results indicate that, given a strong world model, \textbf{FF-JEPA can leverage it to acquire task-specific behaviour in a few-shot manner}.

Apart from data efficiency, in Figure \ref{fig:inference_cost} we also report the inference cost for every approach. FF-JEPA is faster than WM only, since planning towards subgoals is much cheaper using CEM than optimizing the entire trajectory at once. HWM also plans towards subgoals, but uses CEM to find them, causing it to lag behind. Although FF-JEPA closes the gap considerably, Diffusion Policy still has the lowest inference time due to not relying on CEM at all. 

\begin{takeaway}
    The subgoal predictor can learn tasks without requiring actions and in a few-shot manner. Additionally, FF-JEPA achieves lower inference time compared to HWM, a search-based hierarchical world model.
\end{takeaway}

\section{Limitations and future work}

Several limitations of our approach point to promising directions for future work. 
First, although our world-model data carries no task labels, rewards, or success flags, it is not entirely unrelated to the evaluated tasks. PushT and LIBERO-Goal data consist of noisy rollout demonstrations. On Basement, about half of the trajectories pursue goals that include some of the evaluation landmarks. Thus, both the world model and the action prior may benefit from task-directed data coverage. Furthermore, our subgoal predictor is trained on a relatively narrow set of tasks, and generalization to unseen tasks remains unexplored. Training on large-scale datasets with diverse language or multimodal conditioning could enable more general-purpose subgoal prediction.

Second, FF-JEPA's performance depends on both the plannability of the learned representations and the accuracy of the world model. In our experiments, we use an action prior to initialize CEM with actions close to the training distribution. This highlights a limitation of planning with learned world models: predictions may be unreliable when optimization explores out-of-distribution regions. Improving representation plannability and model accuracy, as well as developing more robust mechanisms for constraining planning, could further improve performance. Additionally, planning with CEM makes FF-JEPA 2--5$\times$ slower per step than policies like Diffusion Policy. More efficient and faster search approaches than CEM could benefit FF-JEPA and world model-based planning in general.

Finally, our evaluation is limited to simulated environments and two random seeds. While the results show consistent performance across seeds, evaluating FF-JEPA across more diverse and larger-scale environments would provide a stronger assessment of its scalability and generality. Moreover, Diffusion Policy does not leverage the interaction data available to FF-JEPA, as it is not directly suited to incorporating such data through imitation learning. Future comparisons with policies that can exploit interaction data through relabeling or pretraining would provide a more comprehensive assessment of the benefits of this additional supervision.

\section{Conclusion}
We investigated whether a frozen world model, trained without task labels, can provide the action knowledge needed to solve tasks learned from demonstrations without action labels. Our results suggest that it can. By conditioning a subgoal predictor on the task, FF-JEPA guides the world model towards successive subgoals, enabling long-horizon manipulation and navigation without goal images or high-level search, and achieves performance comparable to policies trained with action supervision. We further suggest that this relies on the world model capturing task-relevant state and progress, while the subgoal predictor can be learned from only a few demonstrations per task. These findings suggest that task and action knowledge can be learned separately, each from the data best suited to it, offering a promising direction towards scaling world-model-based planning to settings where action labels are scarce.

\section*{Acknowledgements}{
    We want to thank Yutong Hu for his earlier involvement. We also appreciate Zehao Wang and Minye Wu for their
feedback during the initial stages of this project. This paper has received funding from the Flemish Government under the Methusalem Funding Scheme (grant agreement n° METH/24/009). The resources and services used in this work were provided by the VSC (Flemish Supercomputer Center), funded by the Research Foundation - Flanders (FWO) and the Flemish Government.
}

\section*{LLM Usage Statement}
We used Claude Code (Anthropic) to implement the code and to orchestrate and run the experiments, following methods and experimental protocols that we defined. All ideas are our own. We drafted the related work and the appendix with Claude Code, starting from works and pointers we provided. Claude Code also generated the code to help create the figures and tables under our direction, following the style and design we specified. We used Claude and ChatGPT (OpenAI) to revise and improve the text, and Claude to help identify inconsistencies and limitations in the paper. The authors reviewed all AI-assisted content and take full responsibility for the paper.

{
    \small
    \bibliographystyle{ieeenat_fullname}
    \bibliography{main}
}

\appendix
\clearpage
\section{The Basement benchmark}
\label{app:basement}

\paragraph{Scene.}
Basement is a 3D model of the basement floor of a university building: two corridors, a hall, a kitchen and a side
corridor, about $47{\times}28$\,m of navigable floor. %

\paragraph{Robot.}
The agent is a simulated Unitree Go2 quadruped with a forward-facing camera. The camera's height (0.395\,m), pitch
($+4^\circ$) and field of view ($78^\circ$) were fitted to footage from a real Go2, and per-channel tone curves and a slight blur match are intended to match its images. The robot is commanded with body-frame velocities $(v_x, v_y, \omega)$ at 5\,Hz, limited to
$v_x \in [-0.4, 1.0]$\,m/s, $|v_y| \le 0.4$\,m/s and $|\omega| \le 1$\,rad/s (an indoor subset of the robot's range). The
commanded velocity is tracked with a first-order lag and per-axis gain, with velocity noise while moving and a small gait
bob of the camera. Collisions are checked against an occupancy map of the same geometry (5\,cm cells; the robot is a
disc of radius 0.25\,m, blocked by any obstacle lower than 0.45\,m, so it can pass under tables). A blocked step leaves the robot in place.

\paragraph{World-model data.}
$\mathcal{D}_{\text{wm}}$ contains 10{,}016 trajectories of 200 frames (40\,s), 2\,M frames in total, from three
behaviours:
\begin{itemize}
  \item \textbf{Goal-seeking} (51\%): pure pursuit along the shortest path to a chain of goals (landmarks or random free
  points 5--30\,m away), at a random speed and with Ornstein--Uhlenbeck action noise of four strengths;
  \item \textbf{Exploration} (30\%): correlated random commands with a forward bias, occasional stops, and turns towards
  open space;
  \item \textbf{Random} (19\%): piecewise-constant random commands without obstacle avoidance.
\end{itemize}
Each episode randomises the camera height, pitch and field of view, image brightness, colour balance, gamma and noise,
and the robot's velocity lag and tracking gain. Starts are spread over the floor. 62\% of the episodes contain at least
one collision, but only 1.6\% of the frames are in collision. The data carries no task labels. We split it into 9{,}014
training and 1{,}002 held-out episodes.

\paragraph{Tasks and demonstrations.}
The 20 tasks are instructions naming a landmark (e.g.\ ``go to the green door in the lobby'', ``go to the TV screen in
the kitchen''), each with a stop pose about 1.2\,m in front of it, facing it. The 2{,}000 demonstrations (100 per task) are
produced by an expert that follows the shortest path to the stop pose with mild action noise, slows down near it, turns to
face the landmark and stops. Starts are drawn over the floor at 3--30\,m of path from the goal (mean 14.8\,m). A
demonstration is kept only if the expert stops at the goal without any collision. Demonstrations last 41--600 steps
(mean 181). We hold out 10 per task (200) and train on 1{,}800.

\paragraph{Evaluation.}
The test set has 300 start poses (15 per task), drawn uniformly over free space at least 0.55\,m from walls, with a
uniformly random heading and 3--30\,m of path to the goal (median 15.3\,m). The expert reaches the goal from all 300.
An episode lasts 600 steps (2 minutes), and the robot succeeds if it is within 0.75\,m and $45^\circ$ of the stop pose at
any step. For goal-image baselines, the goal image is rendered at the stop pose. Two further sets of 100 start poses,
drawn the same way with other seeds, are used for tuning hyperparameters. %
Images are passed through an H.264 encoder with the settings of the dataset videos, as the world model is trained on compressed frames.
\section{Baselines}
\label{app:baselines}

All baselines use the same evaluation starts, step budgets and success criteria as FF-JEPA%
, and two seeds each. The world-model baselines (WM only, HWM) use the same frozen world model and the same action prior as FF-JEPA on each benchmark.

\subsection{Diffusion Policy}
\paragraph{Recipe.} We follow the from-scratch Diffusion Policy baseline of OpenVLA~\cite{kim2024openvla} (see their App.~E),
which uses the DROID implementation~\cite{khazatsky2024droid} of Diffusion Policy~\cite{chi2025dp}. We build it on
LeRobot's~\cite{cadene2026lerobot} conditional 1-D U-Net:
\begin{itemize}
  \item \textbf{Vision:} one ResNet-50~\cite{he2016deep} per camera, ImageNet-initialized, with GroupNorm instead
  of BatchNorm, followed by a linear layer to 512 dimensions (robomimic's VisualCore~\cite{robomimic2021}). Images are resized to $128{\times}128$; training uses colour jitter (0.3) and random $116{\times}116$ crops, testing
  a centre crop.
  \item \textbf{Policy:} observation horizon 2, prediction horizon 16 actions, of which 8 are executed before
  re-planning. U-Net channels $[256, 512, 1024]$, kernel 5, 8 groups, 256-dimensional diffusion-step embedding, FiLM
  conditioning. DDIM~\cite{song2021denoising} with a squared-cosine schedule, 50 training and 10 inference steps,
  $\epsilon$-prediction, 8 noise samples per training item; actions are normalized to $[-1,1]$.
  \item \textbf{Task conditioning:} the instruction is encoded by a frozen DistilBERT~\cite{sanh2020distilbert},
  sum-pooled (768 dimensions), on LIBERO-Goal and Basement. PushT has a single task and no text.
  \item \textbf{Inputs:} the same cameras as FF-JEPA (LIBERO-Goal: agent and wrist view; Basement and PushT: one camera).
  We do not use proprioception to keep comparisons fair, since the other methods also do not use it.
\end{itemize}
The resulting policies have 197\,M (LIBERO-Goal), 142\,M (Basement) and 120\,M (PushT) trainable parameters, plus the
frozen 66\,M-parameter text encoder on the two multi-task benchmarks.

\paragraph{Training.} Each policy is trained on the same task demonstrations as the subgoal predictor (LIBERO-Goal: 378
demonstrations; Basement: 1{,}800; PushT: 1{,}681), but, in contrast to FF-JEPA, also uses their action labels. We train with Adam (learning rate $10^{-4}$, no weight decay, batch size 128; 64 on PushT) in bf16, and evaluate an
exponential moving average of the weights (power 0.75). On LIBERO-Goal and Basement, each policy is trained for 124
epochs (45k steps on LIBERO-Goal, 320k on Basement), and we evaluate the last checkpoint. The reference recipe trains for 25k steps at batch size 128, which amounts to about 62 passes over its LIBERO-Goal training data. Since our datasets differ in size, we use this number of passes (62 epochs) as the minimum training length, with at most twice that (124 epochs), stopping earlier only if the training loss plateaus. No run stopped early, so every policy is trained for 124 epochs (45k steps on LIBERO-Goal, 320k on Basement). On PushT we train for 200k steps, as LeRobot's PushT
configuration, and select the checkpoint (every 5k steps) by the TAC on 256 validation starts that are disjoint from the
test starts; the selected checkpoints (15k and 20k steps) score 0.03--0.18 higher than the last one.

\subsection{WM only}
WM only is the flat planner of HWM~\cite{zhang2026hwm}: CEM towards the encoded goal image with the same world model, the same action prior, and the same CEM settings as FF-JEPA. CEM is optimized for $d/5$ action blocks capped at 20 blocks, where $d$ is the distance (steps) towards the goal. It re-plans after every block in LIBERO-Goal and Basement, and every 5 blocks on PushT. The goal image is the demonstration's last frame on LIBERO-Goal, a render at the goal pose on Basement, and the target pose of the T on PushT.

\subsection{HWM}
\paragraph{Model.} We follow HWM~\cite{zhang2026hwm} and its reference code for PushT: a high-level predictor over learned macro-actions, trained on the frozen world model of each benchmark. Training waypoints follow their App.~B.3: 5 waypoints over a span of 25--70 environment steps with uniformly sampled interior points. A one-layer transformer encodes the actions between two waypoints into a 4-D macro-action. Following HWM, whose high-level predictor is a larger copy of its low-level predictor (DINO-WM's, scaled from 25\,M to 75\,M parameters), our high-level predictor is also a larger copy of our world model's predictor: a causal transformer with depth 10, width 384, MLP 3072, 12 heads and dropout 0.1 (45.5\,M parameters), conditioned on the macro-action instead of the action block. It is trained with an L1 teacher-forcing loss on the latents of the world model's training data. In LIBERO-Goal and Basement, we train with AdamW (learning rate $5{\cdot}10^{-4}$, weight decay 0.01, constant schedule, as DINO-WM's predictor), batch 128, for 20 epochs, and keep the epoch with the lowest validation loss on held-out world-model episodes. %

\paragraph{Planning.} The high level runs CEM over macro-actions with an L1 cost to the goal latent, using the PushT settings of HWM's Table~10 by start--goal distance $d$: 900 samples, 20 iterations, 10 elites and horizon 2 for $d \le 37$; 1{,}500 / 40 / 10 / 4 with a variance EMA of 0.9 for $d \le 62$; and 1{,}200 / 20 / 10 / 5 otherwise, with the horizon growing as $\max(5, \lceil d/15 \rceil)$ up to 7 for longer starts and shrinking as the step budget runs out. The low level plans towards the first predicted waypoint with the same prior-initialized CEM as FF-JEPA (horizon 3 blocks, 15 steps, matching HWM's macro step; cost on the last 2 predicted frames) and re-plans after every block. On PushT we keep 25-step units instead: low-level horizon 5 blocks, re-planning and macro step every 25 steps, high-level horizon up to 5.

\subsection{Goal images on LIBERO-Goal}
LIBERO-Goal's initial states come without a goal image, so the goal-image baselines (WM only, HWM) are evaluated on the 50 held-out demonstrations (5 per task) as full-task windows: start at the demonstration's first frame, goal at its last frame, budget 300 steps. Demonstrations whose final frame does not satisfy LIBERO's task check are excluded (47 remain). Because these results (marked $^\ast$ in \cref{tab:main}) use different start states than the other rows, the exact scores are not strictly comparable, but the large gap remains meaningful: while different start states may shift success rates somewhat, they cannot on their own account for such a big difference between WM only and HWM (4.26\%, 5.32\%) and FF-JEPA (89\%).
\section{World-model probes}
\label{app:probes}

\Cref{sec:wm_analysis} reports the two probes that track success best. Here we define every probe we ran and report all
of them (\cref{tab:probes_full}). All probes use held-out data that neither the world model nor the subgoal predictor
saw, the world models of \cref{sec:wm_analysis} at their selected checkpoints, and both model seeds.

\paragraph{Probes.}
\begin{itemize}
  \item \textbf{Subgoal error.} On 512 held-out windows of the subgoal predictor's own split, we predict the $N{=}3$
  subgoals from the current frame and measure, for each subgoal $k$,
  $\mathbb{E}\|\hat{\mathbf{g}}_k - \mathbf{z}_{t+k\Delta}\|^2 / \mathbb{E}\|\mathbf{z}_t - \mathbf{z}_{t+k\Delta}\|^2$ in
  the world model's raw latent space, i.e.\ the space of the CEM cost. A value of 1 means the predictor is no better
  than copying the current frame. We report the mean over the three subgoals, for the deterministic predictor and for
  one diffusion sample (DDIM 20).
  \item \textbf{Object state beyond the agent.} How much of the objects' state the latent encodes that the agent's own
  position does not explain. We first regress the object state on the agent's position (end effector or pusher, with
  quadratic terms) by least squares, then fit a ridge probe ($\lambda{=}10^{-2}$, standardized features) from the frame
  latent to the residual, and report $R^2$ on held-out episodes (20\%). Targets: the T's angle ($\sin$, $\cos$) and
  position in PushT; the table-plane position of every free object and the position of every articulated joint (drawers,
  stove knob) in LIBERO-Goal, averaged over the objects manipulated in the evaluated episodes. LIBERO-Goal uses 250
  held-out episodes and windows with exact simulator states (11{,}437 frames); PushT uses the held-out expert episodes.
  Basement has no objects that can be manipulated.
  \item \textbf{Monotonicity.} Whether the latent distance tracks task progress, which a goal-distance cost relies on:
  the Spearman correlation between $\|\mathbf{z}_t - \mathbf{z}_T\|^2$ and the remaining steps $T-t$ along each held-out
  demonstration (one frame every 5 steps), averaged over episodes (LIBERO-Goal: 50 demonstrations; Basement: 200
  episodes; PushT: 187 demonstrations).
  \item \textbf{Open-loop error.} From one context frame, the world model rolls out 5 action blocks with the recorded
  actions on 512 held-out windows of its own data. We report
  $\mathbb{E}\|\hat{\mathbf{z}}_5 - \mathbf{z}_5\|^2 / \mathbb{E}\|\mathbf{z} - \bar{\mathbf{z}}\|^2$, where
  $\hat{\mathbf{z}}_5$ is the predicted latent after the fifth block, $\mathbf{z}_5$ the encoded latent of the frame actually reached, $\bar{\mathbf{z}}$ the mean latent over the held-out windows. The expectations are over windows: the prediction error as a fraction of the latent variance (0: perfect; 1: no better than predicting the mean latent). For reference, we give the same error for predicting no change, $\hat{\mathbf{z}}_5 = \mathbf{z}_0$.
  \item \textbf{Action sensitivity.} On 256 held-out windows (3 context frames), the prediction error after one block, relative to predicting no progress, with the expert actions, with zero (no-op) actions, and with the actions of another window. A world model that uses its actions has a much lower error with the expert actions.
  \item \textbf{Object weight in the cost.} How strongly the planning cost reacts to the objects compared with the agent.
  From the same start state we re-render the scene with only the object moved or with only the agent moved by the
  same distance, and take the ratio of the squared latent changes. LIBERO-Goal: 4\,cm object vs 4\,cm end-effector moves
  in the table plane, median over start states. PushT: the T rotated by $15^\circ$ or moved by 20 units vs the pusher
  moved by 20 units (workspace of $512{\times}512$ units), median over 72 expert start states where the pusher and the T
  do not overlap after the move. A value of 1 means an object error costs the planner as much as an equal agent error.
\end{itemize}

\begin{table*}[t]
\centering
\footnotesize
\setlength{\tabcolsep}{3.5pt}
\caption{\textbf{All world-model probes.} Mean over 2 model seeds (largest seed-to-seed std of any probe: 1.09). Success: FF-JEPA Det. / Diff. (PushT: TAC; others: \%). Subgoal err.: $\mathbb{E}\|\hat g_k - z_{t+k\Delta}\|^2 / \mathbb{E}\|z_t - z_{t+k\Delta}\|^2$ in the world model's latent, mean over the 3 subgoals (1 = copying the current frame). Obj.: object state beyond the agent ($R^2$; PushT: the T's angle / position). Mono.: Spearman $\rho$ of the latent distance to the final frame vs the remaining steps. Open-loop:$\mathbb{E}\|\hat{\mathbf{z}}_5 - \mathbf{z}_5\|^2 / \mathbb{E}\|\mathbf{z} - \bar{\mathbf{z}}\|^2$ after 5 predicted steps (copying the first frame in brackets). Action sens.: prediction error after one block relative to no progress, with the expert / no-op / shuffled actions. Obj.\ w.: object weight in the planning cost, $\|\Delta z\|^2$ for a 4\,cm object move over a 4\,cm end-effector move (LIBERO); PushT: the T rotated 15$^\circ$ / moved 20 units over the pusher moved 20 units (workspace 512 units), median over 72 start states. Shaded: selected world model.}
\label{tab:probes_full}
\begin{tabular}{llccccccc}
\toprule
 & & Success & Subgoal err. & Obj. & Mono. & Open-loop & Action sens. & Obj.\ w. \\
 & & ($\uparrow$) & ($\downarrow$) & ($\uparrow$) & ($\uparrow$) & ($\downarrow$) & exp. / no-op / shuf. & \\
\midrule
PushT & Inverse dyn. & .48 / .57 & 0.06 / 0.10 & 0.63 / 0.91 & 0.70 & 0.05 (1.36) & 0.11 / 0.84 / 0.18 & 2.8 / 2.9 \\
 & SIGReg & \best .82 / .76 & \best 0.07 / 0.11 & \best 0.97 / 0.97 & \best 0.33 & \best 0.02 (1.43) & \best 0.01 / 0.84 / 0.05 & \best 12.9 / 13.3 \\
 & SpecWM & .71 / .75 & 0.05 / 0.08 & 0.54 / 0.84 & 0.80 & 0.05 (1.61) & 0.02 / 0.82 / 0.06 & 4.7 / 4.5 \\
 & DINO-WM & .53 / .49 & 0.16 / 0.52 & 0.78 / 0.88 & 0.79 & 0.18 (1.27) & 0.19 / 0.88 / 0.27 & 4.1 / 2.7 \\
\midrule
LIBERO-Goal & Inverse dyn. & \best 89.0 / 82.5 & \best 0.28 / 0.30 & \best 0.57 & \best 0.67 & \best 0.11 (0.66) & \best 0.24 / 2.15 / 1.28 & \best 0.10 \\
 & SIGReg & 6.8 / 3.5 & 0.48 / 0.63 & $-$0.61 & 0.67 & 0.13 (0.59) & 0.30 / 0.90 / 1.27 & 0.01 \\
 & SpecWM & 66.0 / 60.8 & 0.30 / 0.37 & 0.42 & 0.91 & 0.16 (0.98) & 0.22 / 1.13 / 1.23 & 0.30 \\
 & DINO-WM & 81.5 / 10.8 & 0.55 / 0.82 & 0.45 & 0.85 & 0.54 (1.37) & 0.44 / 0.71 / 0.80 & 0.17 \\
\midrule
Basement & Inverse dyn. & 58.2 / 59.5 & 0.10 / 0.43 & n/a & 0.47 & 0.16 (1.73) & 0.08 / 0.58 / 1.09 & n/a \\
 & SIGReg & 38.7 / 45.7 & 0.03 / 0.18 & n/a & 0.63 & 0.24 (1.08) & 0.20 / 0.57 / 0.96 & n/a \\
 & SpecWM & \best 82.0 / 90.7 & \best 0.16 / 0.37 & \best n/a & \best 0.80 & \best 0.09 (1.14) & \best 0.09 / 0.48 / 0.99 & \best n/a \\
 & DINO-WM & 59.2 / 49.7 & 0.34 / 0.90 & n/a & 0.43 & 0.71 (1.79) & 0.31 / 0.68 / 0.81 & n/a \\
\bottomrule
\end{tabular}
\end{table*}

\paragraph{Findings.}
\begin{itemize}
  \item \textbf{Prediction accuracy and action use do not order the world models.} Every world model uses its actions on
  every benchmark. The expert-action error is 1.6--100$\times$ lower than the no-op or shuffled error. DINO-WM and, on PushT,
  inverse dynamics are the least sensitive. Open-loop error orders them neither by success nor consistently across benchmarks: DINO-WM predicts
  worst everywhere yet is the second-best world model on LIBERO-Goal with the deterministic predictor.
  \item \textbf{Navigation follows monotonicity.} On Basement, SpecWM's latent distance tracks progress best
  ($\rho{=}0.80$) and SpecWM is the best world model by a wide margin. It also has the lowest open-loop error there.
  Monotonicity identifies the winner but does not order the rest: SIGReg is second in monotonicity and last in
  success. The subgoal error does not explain Basement either: SIGReg's latent is the easiest to predict (0.03).
  \item \textbf{Manipulation needs object information.} On LIBERO-Goal, the object state beyond the arm orders the
  four world models exactly as their success with the deterministic predictor, and SIGReg, whose latent carries no
  object information beyond the arm ($R^2{=}{-}0.61$), collapses. On PushT, the best world model, SIGReg, encodes the
  T's angle far better than the others ($R^2$ 0.97 vs at most 0.78) and gives it by far the largest weight in the cost.
  \item \textbf{Monotonicity is not needed for manipulation.} The best world models on PushT and LIBERO-Goal have the
  least monotonic latents ($\rho$ 0.33 and 0.67), while SpecWM is the most monotonic on both and is not the best on
  either. A likely reason is that the subgoal predictor supplies the long-range direction, so CEM only needs a
  latent distance that is correct locally, towards the next subgoal.
  \item \textbf{The subgoal predictor decides between predictors.} For a fixed world model, the predictor whose subgoals
  are harder to predict plans worse. The clearest case is DINO-WM on LIBERO-Goal: its deterministic predictor reaches
  81.5\% success with a subgoal error of 0.55, while its diffusion predictor, at 0.82, barely beats copying the current
  frame and reaches 10.8\%.
\end{itemize}

\paragraph{Object information or object weight?}
On LIBERO-Goal we tested whether SpecWM falls behind inverse dynamics because its planning cost ignores objects. It does not: SpecWM gives objects the largest weight in the cost of all world models (0.30 vs 0.10 for inverse dynamics), and along held-out demonstrations its latent goal distance follows the objects' distance to their goal state more than any other world model's does (partial rank correlation controlling for the end effector: 0.35 vs 0.22--0.28). On LIBERO-Goal, success therefore requires that the object state can be read precisely from the latent, not how much the cost weighs it. On PushT, where the T dominates the image, every world model except inverse dynamics weighs the T more than the pusher, and the weight orders the world models, while the linear decodability of the T's pose does not. In both cases, the best world model is the one that represents the manipulated object best, in the form the planning cost can use. %

\paragraph{Caveats.}
The probes are correlational and cover four world models per benchmark, so they support but do not prove these explanations. On LIBERO-Goal, the cream cheese is manipulated in few held-out windows and its per-object $R^2$ is negative for every world model. For DINO-WM, the object probe uses the patch tokens averaged per view. %

\section{Qualitative results}
\label{app:qualitative}

\begin{table}[ht]
\centering
\small\setlength{\tabcolsep}{3pt}
\caption{\textbf{Getting stuck and getting free on Basement} (\%). A robot is considered stuck if it moved $<0.2$\,m over 5 re-plans while $\ge 0.75$\,m from the goal. Re-run with logging enabled. Tested on 300 windows.}
\label{tab:basement_stuck}
\begin{tabular}{lcc}
\toprule
 & FF-JEPA Det. & FF-JEPA Diff. \\
\midrule
Success & 81.3\sd{4.7} & \textbf{90.8}\sd{0.2} \\
Gets stuck at some point & 26.5\sd{2.1} & \textbf{21.5}\sd{0.2} \\
\quad of which frees itself & 46.5\sd{13.5} & \textbf{77.5}\sd{0.8} \\
\midrule
Failures: stuck at the end & 87.5\sd{6.9} & 74.5\sd{5.8} \\
Failures: freed, but out of time & 11.7\sd{8.0} & 25.5\sd{5.8} \\
\bottomrule
\end{tabular}
\end{table}

\begin{figure*}[t]
\centering
\includegraphics[width=\textwidth]{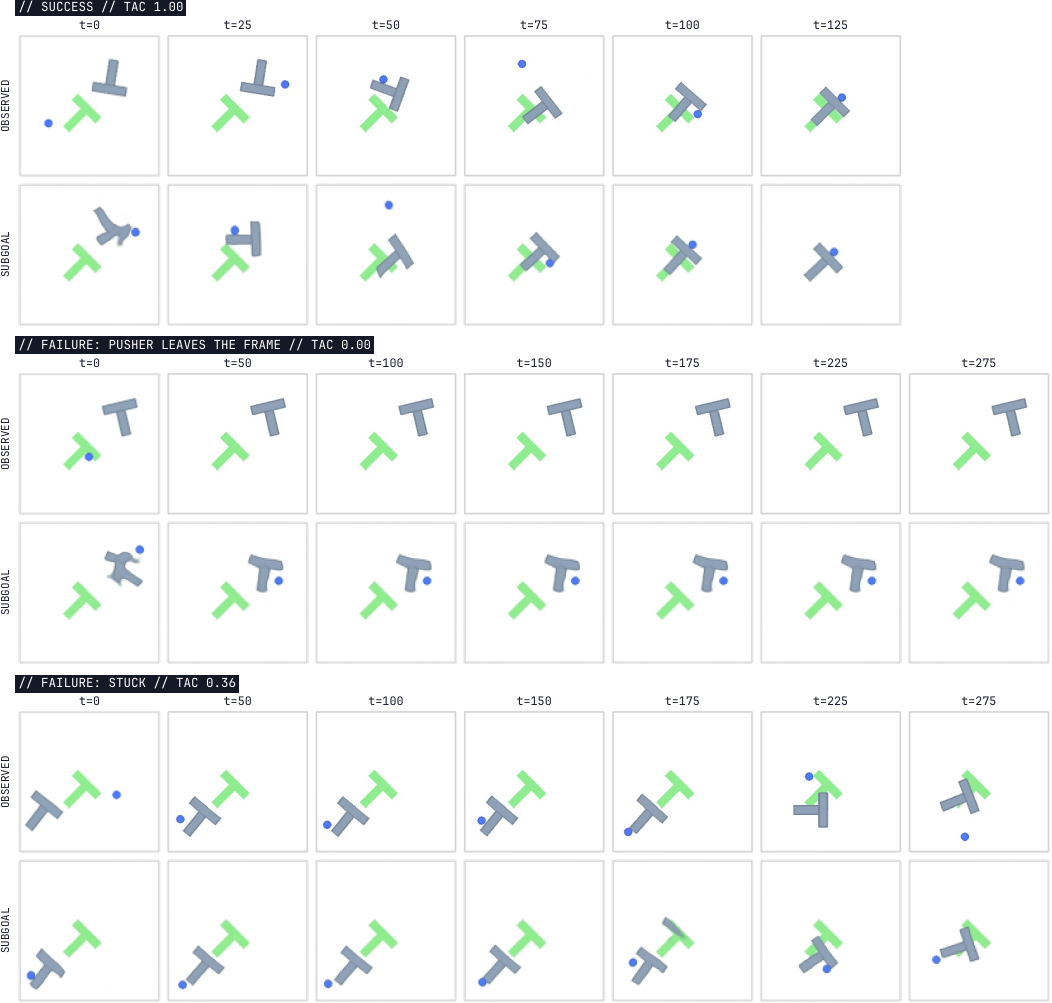}
\caption{\textbf{PushT rollouts} (re-plan every 25 steps). Top row of each example: observed frame; bottom: decoded
first subgoal. \emph{Success}: the subgoals rotate and push the T onto the target (95\% coverage at step 150).
\emph{Pusher leaves the image} (the most frequent failure): the pusher exits within the first re-plans, and the subgoals
keep imagining it beside the T for the rest of the episode. \emph{Stuck}: the pusher stays at a corner of the T for
about 150 steps with nearly identical subgoals, and moves the T too late (TAC 0.36). Rendered from a re-run of the
evaluation with recording enabled.}
\label{fig:qual_pusht}
\end{figure*}

\begin{figure*}[t]
\centering
\includegraphics[width=\textwidth]{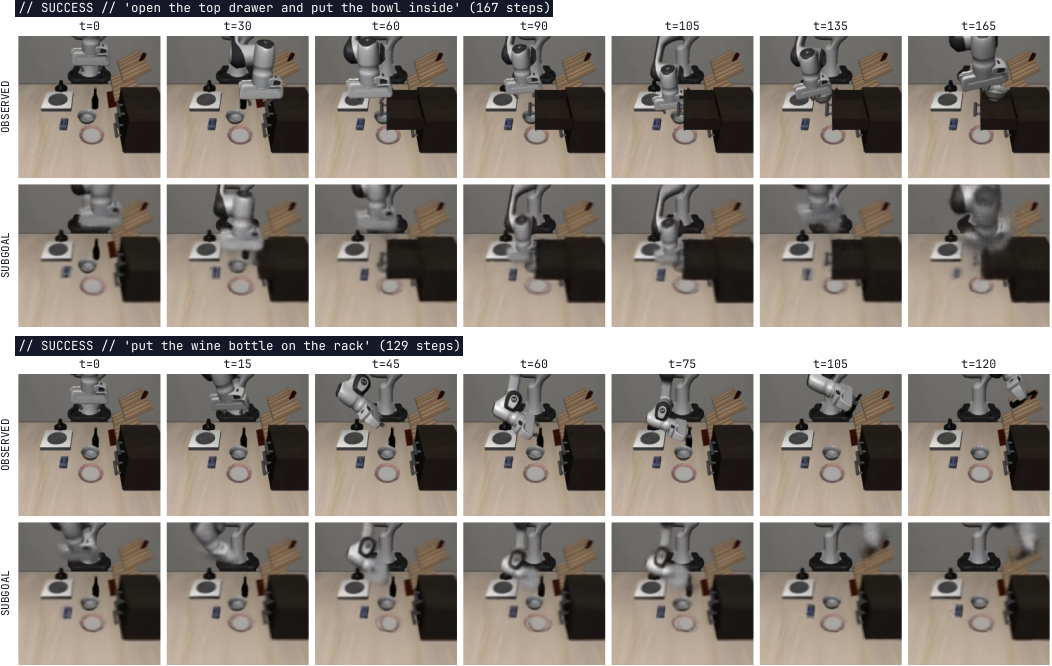}
\caption{\textbf{LIBERO-Goal, successful rollouts.} Top: observed agent view at each re-plan (every 15 steps); bottom:
decoded first subgoal predicted at that re-plan. The subgoals lead the arm through the stages of each task: opening the
drawer, grasping the bowl and placing it inside (top), and grasping the wine bottle and placing it on the rack (bottom).}
\label{fig:qual_libero_success}
\end{figure*}

\begin{figure*}[t]
\centering
\includegraphics[width=\textwidth]{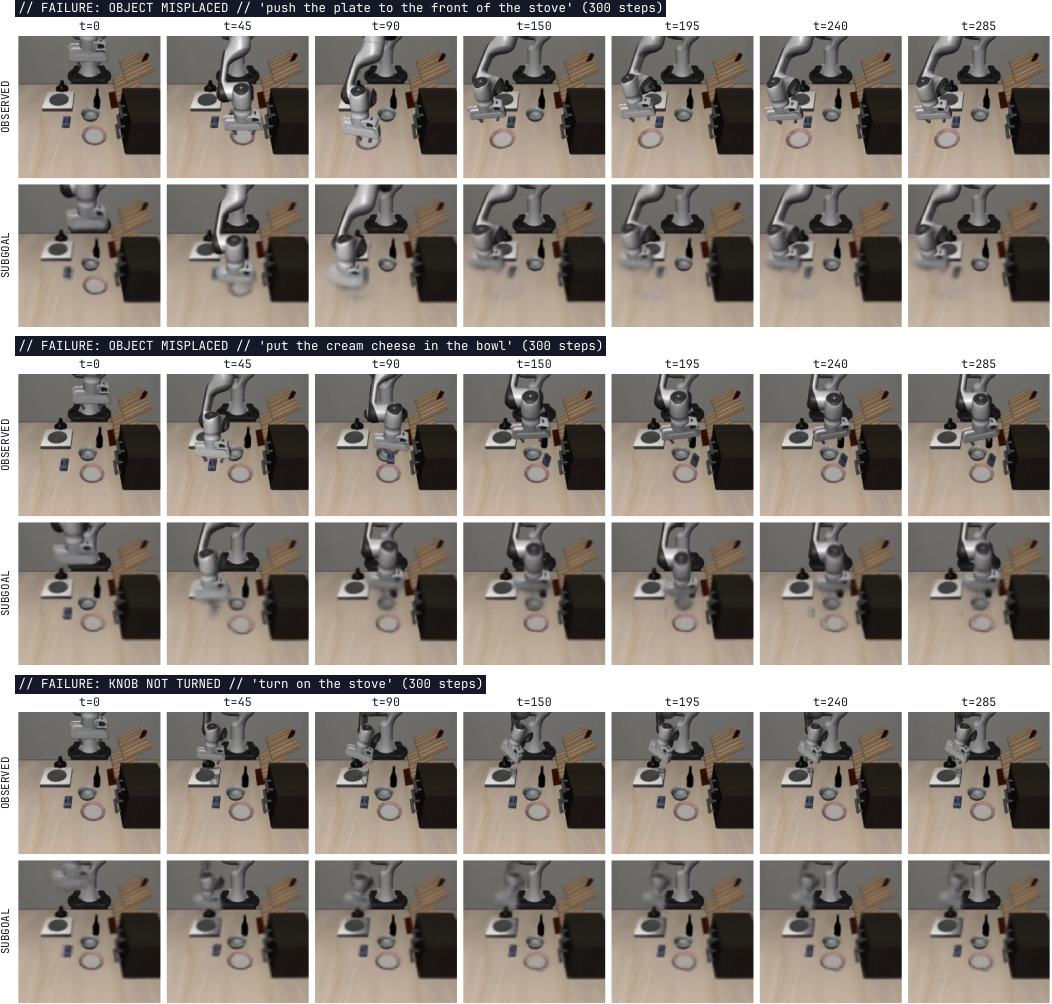}
\caption{\textbf{LIBERO-Goal, failed rollouts} (the three most frequent failure types). \emph{Plate}: the plate is pushed
about 20\,cm but stops short of the target region; later subgoals show the arm near the stove with the plate barely
visible. \emph{Cream cheese}: the cream cheese is carried but released next to the bowl, while the subgoals keep showing
the arm above the bowl. \emph{Stove}: the arm reaches the stove, but the subgoals stay blurred around the knob and the
knob is never turned. All three episodes time out at 300 steps.}
\label{fig:qual_libero_failure}
\end{figure*}

\begin{figure*}[t]
\centering
\includegraphics[width=\textwidth]{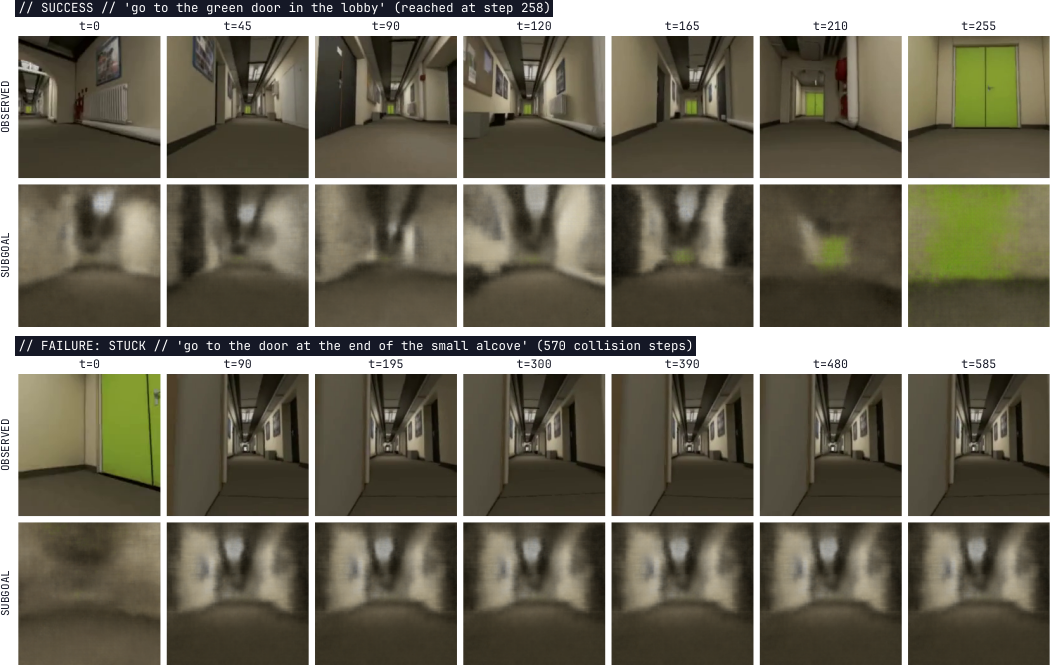}
\caption{\textbf{Basement rollouts.} Top: robot camera at each re-plan (every 15 steps); bottom: decoded first subgoal.
\emph{Success}: from a start about 30\,m away, the subgoals follow the corridor and turn into the green door as the robot
approaches it (goal reached at step 258). \emph{Failure} (the most frequent type): after a turn the robot is pinned
against the wall of a side corridor; the deterministic predictor issues the same subgoal at every re-plan and the robot
stays stuck for the rest of the episode (570 collision steps).}
\label{fig:qual_basement}
\end{figure*}

\Cref{fig:qual_pusht,fig:qual_libero_success,fig:qual_libero_failure,fig:qual_basement} show rollouts of FF-JEPA with the deterministic
subgoal predictor (seed 1). Each column is a re-plan: the top row is the camera image at that step, the bottom row the
first subgoal predicted from it, decoded to an image by a convolutional decoder trained afterwards on the frozen world
model. The decoder only serves visualisation; planning never decodes. The episodes are re-runs of the evaluation protocol
with videos enabled; the failures were chosen to cover the most frequent failure types below.

\paragraph{How FF-JEPA fails.}
\begin{itemize}
  \item \textbf{PushT.} The dominant failure is the pusher leaving the image. It happens in 29\% of the episodes of the
  deterministic predictor (150 of 512, both seeds), and these episodes account for 49 of the 50 that end with almost no
  coverage and for about two thirds of the TAC lost. Once the pusher is outside the image, the latent no longer contains
  it, and the predicted subgoals keep showing it next to the T (\cref{fig:qual_pusht}, middle), so the agent does not
  recover. Less often, the pusher pins itself against a corner of the T, where the subgoals barely change for
  several re-plans (\cref{fig:qual_pusht}, bottom).
  \item \textbf{LIBERO-Goal.} Of the 23 failed episodes of the deterministic predictor (seed 1, 88.5\% success), 14 moved
  the target object but did not place it where the instruction asks (the plate pushed short of the region in front of
  the stove, the cream cheese released next to the bowl), 5 never moved it (grasp or approach failed), and 4 touched but
  did not turn the stove knob. Failures therefore concentrate in the contact phase, where success depends on a precise
  object pose. In the examples of \cref{fig:qual_libero_failure}, the decoded subgoals reproduce the arm and the scene
  sharply, while the manipulated object appears faint or blurred in the later subgoals, consistent with the limited
  object information of the latent (\cref{app:probes}). LIBERO episodes end at success, so every failure runs the full
  300 steps.
  \item \textbf{Basement.} About 88\% of the failures of the deterministic predictor end with the robot stuck against an
  obstacle and 11\% freed themselves but ran out of time (\cref{tab:basement_stuck}). In the stuck
  case, the deterministic predictor issues the same subgoal at every re-plan (\cref{fig:qual_basement}, bottom), because
  the world model predicts progress where the robot is blocked (\cref{sec:detsto}). %
\end{itemize}

\section{Datasets}
\label{app:datasets}

\paragraph{PushT.}
We use the PushT dataset released with LeWM~\cite{maes2026lewm}: 18{,}685 noisy expert episodes. The 10\% of episodes
that reach the highest coverage of the target (best coverage $\ge 88.2\%$), each cut at its best step, form the task
demonstrations (1{,}868: 1{,}681 for training, 187 held out); the remaining 16{,}817 episodes form $\mathcal{D}_{\text{wm}}$
(15{,}135 for training, 1{,}682 held out). The two sets are therefore disjoint, and the world model never sees the best demonstrations. They also differ sharply in quality: $\mathcal{D}_{\text{task}}$ is clearly expert, with a mean TAC score of 0.957, while $\mathcal{D}_{\text{wm}}$ is much noisier and suboptimal, with a mean TAC of only 0.636.

\paragraph{LIBERO-Goal.}
$\mathcal{D}_{\text{wm}}$ consists of 20{,}000 rollouts of SmolVLA~\cite{shukor2025smolvla} on the 10 LIBERO-Goal tasks
(3.6\,M frames at 20\,Hz; agent and wrist cameras at $256{\times}256$, H.264-compressed). To diversify the data, each
rollout adds Gaussian noise to the policy's actions, with a standard deviation drawn per episode from
$\{0, 0, 0.1, 0.2, 0.35\}$. 62.6\% of the rollouts succeed. We do not use their task labels or success flags. The world
model is trained on 18{,}000 rollouts and 2{,}000 are held out. The task demonstrations are LIBERO's human
demonstrations of the goal suite~\cite{liu2023libero} (428); 5 per task (50) are held out at random and the remaining 378
train the subgoal predictor and Diffusion Policy.

\paragraph{Basement.} See App.~\ref{app:basement}.

\section{Does planning or the prior do the work?}
\label{app:abl3}

The prior initializes CEM%
. To separate the contributions of the prior and of the search, we compare
FF-JEPA (deterministic predictor) with two variants that keep everything else fixed:
\begin{itemize}
  \item \textbf{Plain CEM:} CEM starts from the standard proposal $\mathcal{N}(0, I)$ in z-scored action space, without
  the prior.
  \item \textbf{Prior only:} no search and no subgoals. The agent executes the prior's mean plan, obtained by rolling the
  prior out through the world model over the planning horizon, and re-plans as FF-JEPA does. It sees neither the task
  nor a goal.
\end{itemize}

\begin{table}[h]
\centering
\footnotesize
\setlength{\tabcolsep}{3pt}
\caption{\textbf{Planning vs the action prior.} FF-JEPA Det., selected world model; mean \sd{std} over 2 model seeds.}
\label{tab:abl3}
\begin{tabular}{@{}lccc@{}}
\toprule
 & PushT & LIBERO-Goal & Basement \\
 & {\scriptsize TAC} & {\scriptsize success (\%)} & {\scriptsize success (\%)} \\
\midrule
Prior only & 0.25\sd{0.01} & \phantom{0}3.75\sd{2.47} & \phantom{0}6.33\sd{1.41} \\
Plain CEM & 0.79\sd{0.08} & 67.75\sd{8.84} & 75.17\sd{2.59} \\
Prior-init.\ CEM (FF-JEPA) & \textbf{0.82}\sd{0.01} & \textbf{89.00}\sd{0.71} & \textbf{82.00}\sd{2.36} \\
\bottomrule
\end{tabular}
\end{table}

The prior alone solves almost nothing: it is task-blind and only proposes plausible actions. 
Planning towards the predicted subgoals does the work: plain CEM already reaches 0.79 TAC on PushT, 67.8\% on LIBERO-Goal and 75.2\% on Basement. Starting CEM from the prior adds 21 points on LIBERO-Goal and 7 on Basement. For PushT the difference is not big enough to be conclusive, but we keep using the prior to stay consistent with the other benchmarks.

\end{document}